\documentclass[conference]{IEEEtran}
\IEEEoverridecommandlockouts
\usepackage{cite}
\usepackage{amsmath,amssymb,amsfonts}
\usepackage{graphicx}
\usepackage{textcomp}
\usepackage{xcolor}
\usepackage{booktabs} % For formal tables
\usepackage{tabularx}
\usepackage{color}
\usepackage{hyperref}
\usepackage{xargs}
\usepackage{algorithm}
\usepackage[noend]{algpseudocode}
\usepackage{listings}
\usepackage{multirow}
\usepackage{colortbl}
\usepackage{hhline}
\usepackage{xspace}
\usepackage{tikz}
\usepackage{array}
\usepackage[outline]{contour}
\usepackage{blindtext}
\usepackage{subcaption}
\usepackage{arydshln}
\usepackage{float}
\usepackage{makecell}

\contourlength{0.3em}
\contournumber{250}
\usetikzlibrary{arrows.meta, positioning, arrows, shapes, shapes.geometric, 
patterns, decorations, calc, positioning}

\newenvironment{conditions}
  {\par\vspace{\abovedisplayskip}\noindent\begin{tabular}{>{$}l<{$} @{${}={}$} l}}
  {\end{tabular}\par\vspace{\belowdisplayskip}}
\def\BibTeX{{\rm B\kern-.05em{\sc i\kern-.025em b}\kern-.08em
    T\kern-.1667em\lower.7ex\hbox{E}\kern-.125emX}}
\begin{document}

\newcommand{\shorttitle}{SUDS}
\newcommand{\SUDS}{\textit{SUDS}}
\newcommand{\metric}{\textbf{HADAM}}
\newcommand{\longmetric}{Harmonized Annotated Data Accuracy Metric}

\title{SUDS: A Strategy for Unsupervised Drift Sampling\\
\iffalse
{\footnotesize \textsuperscript{*}Note: Sub-titles are not captured in Xplore and
should not be used}
\thanks{Identify applicable funding agency here. If none, delete this.}
\fi
}
\author{\IEEEauthorblockN{Christofer Fellicious}
\IEEEauthorblockA{%\textit{Chair of Data Science} \\
\textit{University of Passau}\\
%City, Country \\
christofer.fellicious@uni-passau.de}
\and
\IEEEauthorblockN{Lorenz Wendlinger}
\IEEEauthorblockA{%\textit{dept. name of organization (of Aff.)} \\
\textit{University of Passau}\\
%City, Country \\
lorenz.wendlinger@uni-passau.de}
\and
\IEEEauthorblockN{Mario Gancarski}
\IEEEauthorblockA{%\textit{dept. name of organization (of Aff.)} \\
\textit{University of Passau, INSA Lyon}\\
%City, Country \\
mario.gancarski@insa-lyon.fr}
\and
\IEEEauthorblockN{Jelena Mitrovi\'{c}}
\IEEEauthorblockA{%\textit{dept. name of organization (of Aff.)} \\
\textit{University of Passau},\\
The Institute for Artificial Intelligence Research and Development of Serbia \\
jelena.mitrovic@uni-passau.de}
\and
\IEEEauthorblockN{Michael Granitzer}
\IEEEauthorblockA{%\textit{dept. name of organization (of Aff.)} \\
\textit{University of Passau}\\
%City, Country \\
michael.granitzer@uni-passau.de}
}
\maketitle

\begin{abstract}
Supervised machine learning often encounters concept drift, where the data distribution changes over time, degrading model performance. Existing drift detection methods focus on identifying these shifts but often overlook the challenge of acquiring labeled data for model retraining after a shift occurs. We present the Strategy for Drift Sampling~(\SUDS), a novel method that selects homogeneous samples for retraining using existing drift detection algorithms, thereby enhancing model adaptability to evolving data.~\SUDS~seamlessly integrates with current drift detection techniques. We also introduce the Harmonized Annotated Data Accuracy Metric~(HADAM), a metric that evaluates classifier performance in relation to the quantity of annotated data required to achieve the stated performance, thereby taking into account the difficulty of acquiring labeled data.

Our contributions are twofold:~\SUDS~combines drift detection with strategic sampling to improve the retraining process, and~\metric~provides a metric that balances classifier performance with the amount of labeled data, ensuring efficient resource utilization. Empirical results demonstrate the efficacy of~\SUDS~in optimizing labeled data use in dynamic environments, significantly improving the performance of machine learning applications in real-world scenarios.
Our code is open source and available at~\url{https://github.com/cfellicious/SUDS/}
\end{abstract}


\section{Introduction}
\label{sec:introduction}
In classical machine learning, most models train on a static dataset with the underlying assumption that the data distribution is static. 
This assumption works for speech detection, optical character recognition, and so forth, where little to no changes occur over the years.
However, many domains such as credit card fraud detection~\cite{ghosh2022nag}, energy consumption~\cite{jagait2021load}, production forecasting~\cite{xue2021data}, and many others are very dynamic and do not adhere to the static distribution model. 
These domains are examples where the underlying data distribution changes over time and this is known as concept drift.
\v{Z}liobait{\.e} et al.~\cite{vzliobaite2016overview} explain the different application domains of concept drift in detail.

Concept drift alters data distribution, impacting model performance and necessitating model updates. However, obtaining labeled data for retraining can be costly in terms of time, effort, and resources, often requiring manual expert labeling. 
For instance, in credit card fraud detection, delays in acquiring labeled data may justify the use of semi-supervised methods that exploit future unlabeled transactions to improve fraud detection~\cite{dastidar2022importance}. 
Additionally, the scarcity of positive instances further complicates the labeling process~\cite{eng2003sample}.
Therefore, it is crucial to adapt to distribution changes with minimal training data. While algorithmic performance typically garners significant attention, the need for annotated ground truth data often remains overlooked. Sample size estimation, a common practice in fields like medicine and economics, highlights the expense and challenge of obtaining ground truth data~\cite{eng2003sample}. Real-world datasets typically exhibit gradual rather than abrupt drifts, which can contaminate the model's training data when drift is detected in between distribution changes.

Our primary contribution in this paper is~\SUDS, a method that piggybacks on existing unsupervised drift detection methods to select homogeneous data points for retraining a model after the occurrence of a drift.
~\SUDS~operates on the idea, that we can benefit from an analysis of the data present when a drift is detected, and exploit the most recent data to produce a more homogeneous collection of data points for retraining the classifier. 
Using \SUDS, we demonstrate that our method can outperform existing algorithms while requiring only 20\% of the training data compared to these algorithms.
Our secondary contribution is a performance metric, the~\longmetric~(\metric),~that focuses on the performance of a model along with the amount of annotated samples required to achieve that performance. 
With~\metric,~\SUDS~outperforms its counterparts in most datasets, showing excellent performance particularly in real-world datasets.

We organize the paper as follows: we introduce the problem and research statement in~\autoref{sec:introduction} with the related work explained~\autoref{sec:related_work}. Our proposed approach is presented in~\autoref{sec:method} with~\autoref{sec:results} containing the results of our experiments. We then present our conclusions in~\autoref{sec:conclusion}. We provide an extended version of this paper along with the hyperparameter experiments in the supplementary material.
\section{Related Work}
\label{sec:related_work}
%We can categorize drift detection methods into supervised and unsupervised methods based on whether the target labels are available. Supervised methods have feature labels available to them and can be used to detect drift with the additional information they provide. On the other hand, unsupervised drifts cannot access the feature labels and only have to base their predictions by examining the feature vectors.

%The Drift Detection Method(DDM) proposed by Gama et al.~\cite{Gama2004} is one of the earliest works in drift detection. The method is a supervised method that calculates the error rate of the trained classifier and signals a drift warning and drift detected based on the error rate thresholds. The theory behind this method is that the error rate of the classifier increases w.r.t. the change in the underlying data distribution. By monitoring this error rate, the algorithm detects drifts.
We look at some of the unsupervised methods in concept drift detection as it pertains more towards this work.
Sethi et al.~\cite{Sethia5090} proposed the Margin Density Drift Detection(MD3) method for detecting concept drifts from unlabelled data streams. "The proposed methodology uses the number of samples in a classifier's region of uncertainty as a metric for detecting drift."
%Dos Reis et al.~\cite{Reis5090} implement the Incremental Kolmogorov-Smirnov algorithm that performs a test with the Kolmogorov-Smirnov hypothesis using two sample sets that change over time. It can either be an insertion or a removal of a sample observation. The algorithm employs a randomized tree that performs the operations with an O(logN) complexity (where N is the number of sample observations) with high probability and calculates the Kolmogorov-Smirnov test in O(1) complexity. The Incremental Kolmogorov-Smirnov test detects concept drifts without true labels. 
Koh~\cite{Sing5090} introduces a distinction between two types of drifts: local and global drifts. 
Koh's contribution relies on two different drift detection methods based on the two different drift types. The contributions are using fixed windows for their purpose.
Mustafa et al.~\cite{Mustafa5090} proposes a denoising autoencoder to rebuild an input vector from a corrupted version and a nonparametric multidimensional change point detection method. The method works by maintaining several ensembles using KNN models and testing whether the new instances belong to one of the ensemble's classifiers or are outliers. If the new instance is an outlier, it is stored in a frequently checked buffer to verify if the algorithm should add a new classifier to the ensemble.
The One-Class Drift Detector proposed by Gözüaçik et al.~\cite{Goezueacik2020} consists of a sliding window of the DataStream to which we append a sliding window that they called "outlier's window." This window gives the result of the classification of each instance by a One Class SVM trained on a current data set. A drift is detected whenever the outlier's rate exceeds a threshold parameter.
We can take the following proposed methods as examples of the first type: 
Costa et al.~\cite{Institute5090} developed the Drift Detection Method based on Active Learning~(DDAL)~that uses active learning to detect changes in data streams.
We divide the data stream into fixed-size batches and train a model on the initial batch. We then predict the labels of subsequent batches.
The algorithm computes the uncertainties, updating the minimum and maximum values. The algorithm signals a drift if the updated difference between min and max is higher than a threshold.
The algorithm updates the model using the newly labelled instances in the current batch.
Liu et al.~\cite{CCBYNCND} proposed a method based on the  Nearest Neighbor-based Density Variation Identification algorithm.
It relies on a k-nearest neighbor-based space partitioning schema, 
a distance function over the density discrepancies, and a statistically significant test to perform the detection.
Gözüaçik et al.\cite{Omer2019} proposed the Discriminative Drift Detector(D3). The algorithm works by accumulating the samples into two windows that are not the same size.
The algorithm then trains a classifier on both windows, with each window getting its label. 
If the classifier can distinguish between the two windows with a certain threshold, then the algorithm signals a concept drift. The underlying assumption is that if both windows contain data points of the same underlying distribution, the classifier will fail to assign labels correctly, and the new window will contain samples that belong to the drifted distribution.
Gulcan and Can proposed the Label Dependency Drift Detector(LD3) method in multi-label classification. The authors use a label ranking method and exploit the temporal dependencies between labels using this ranking. The method exploits correlations between labels in multi-label data streams. The algorithm considers two windows, one for the old and the other for the new samples. Once these windows are full, the algorithm generates co-occurrence matrices. We obtain these matrices by counting the occurrences of each class label. Each label is then ranked based on the frequencies. The similarity between rankings provides an idea of whether a concept drift occurred.
%Jain et al. proposed a method to detect concept drifts in network data~\cite{jain2022k}. The authors propose two different methods for concept drift detection combined with a Support Vector Machine~(SVM)~classifier for anomaly detection. The authors propose "Error Rate Based Concept Drift Detection" and "Data Distribution based Concept Drift Detection" for Concept Drift Detection. The proposed method contains a hybrid Concept Drift Detection~(CDD)~framework to identify drifts in network traffic. The severity of the drifts is measured using KL-Divergence, Kappa statistics, and accuracy metrics. Then, the authors use a K-Means-based clustering method to reduce the sample size of captured network traffic and training set updation.
In recent years neural networks and Generative Adversarial Networks are used to detect concept drifts\cite{fellicious2022neural, fellicious2024driftgan}. Especially Fellicious et al. uses a GAN to identify historical drifts and use the past data to increase the training data size. This allows for better performance of the model and less susceptible to noise.
Cerquiera et al. proposed a method using the student-teacher model~\cite{cerqueira2023studd}. The method uses the teacher model to learn and the output of the teacher model is learnt by the student. For incoming instances the difference is then computed and based on the change in the output of the student a drift is signalled. 
\section{Method}
\label{sec:method}
In many machine learning algorithms, the data available at the point of drift detection could be from different data distributions~\cite{Survey2013}. 
This heterogeneous data is because data distribution changes are usually gradual, and the drift detection point may not precisely align with the actual point of drift as drift detection depends on specific hyperparameters used by the detection algorithm. 
Training on such heterogeneous data can reduce the classifier's efficacy and negatively impact future drift detection by poisoning the training set. 
To address this issue, we propose a method that leverages the existing classifier to select a more homogeneous subset of data for retraining. 
Our approach aims to enhance overall model performance by ensuring that the training data is homogeneous in its distribution, thereby improving both the retraining process and future drift detection accuracy.

In our work, we introduce terms that we use throughout this work. We refer to the current distribution or~$D_{curr}$~as the distribution on which the model was trained. A concept drift occurs when the current distribution~($D_{curr}$)~changes into a newer distribution. We refer to this changed distribution as the newer distribution or~$D_{next}$. Once the model identifies a concept drift from $D_{curr}$, we retrain the model on~$D_{next}$. Also, we use $X$ to denote input feature vectors and $W$ to denote a window where we consider the input vectors within that window.

Real-world observations from different domains show that concept drift is usually a gradual drift where the data distribution changes over time as in~\autoref{eqn:gradual_drift}~\cite{Survey2013}.
\begin{equation}
    X_t\in D_{curr} \wedge X_{t+1...k}\in\{D_{curr}, D_{next}\} \wedge X_{k+1...n}\in D_{next}
    \label{eqn:gradual_drift}
\end{equation}
where, $X_t$ is the input features at time t,~$D_{curr}$~being the current distribution under consideration and~$D_{next}$~being the changed distribution after the concept drift.
As for our method, we also do not expect a sudden change in the data distribution~(abrupt drift)~as given in~\autoref{eqn:abrupt_drift}.
\begin{equation}
    X_t \in D_{curr} \wedge X_{t+1...n} \notin D_{curr}
    \label{eqn:abrupt_drift}
\end{equation}

In the case of a non-abrupt drift, the samples used for training could belong to a heterogeneous data distribution as shown in~\autoref{eqn:window_distributions}:
\begin{equation}
    W_{k...k+n}\in \{D_{curr},D_{next}\}
    \label{eqn:window_distributions}
\end{equation}
where $W_{k..k+n}$ is the selected training window for the model.

From \autoref{eqn:window_distributions}, we know that the training window comprises a heterogeneous distribution instead of the optimal homogeneous distribution due to the tolerance and threshold hyperparameters of the corresponding drift detection algorithm. 
In such scenarios, training the model on mixed distributions can degrade its performance, in both label prediction and drift detection. 
To address this issue, one solution is to discard an arbitrary number of samples before collecting data to retrain the model.
However, such a waiting strategy could be problematic as it delays model updates and we need to be sure that the next concept drift is sufficiently far into the future.
Therefore, a strategy that selects only the homogeneous data points belonging to the new distribution~($D_{next}$) could provide significant advantages.
This approach should ensure better quality training data for the model, potentially improving the prediction accuracy while lowering the labeling costs in time and human effort.

Our method~"Strategy for Unsupervised Drift Sampling~(\SUDS)"~builds on the assumption that a more homogeneous training data provides better drift detection and performance. 
Our method works by selecting a window around the point of detected drift occurrence and use a classifier to identify the data points that only belong to the newer distribution.
We do this by using the drift detection model itself to detect out of distribution samples w.r.t. the current data distribution and rejecting them.
This is given in~\autoref{eqn:model_retraining}. 
\begin{equation}
    \{W_{train}\in X | \forall X \in D_{next}\}
    \label{eqn:model_retraining}
\end{equation}
where $W_{train}$ refers to the samples for retraining the model and $D_{next}$ refers to the new data distribution. This approach assumes that the majority of data points in $D_{next}$ belong to the new class which allows us to train a selector for similar examples in the distribution prior to the shift. 

As our method essentially treats the drift detector as a black-box model and only modifies the training data selection, it is compatible with all established drift detection techniques.
%Our method builds upon existing approaches, necessitating its integration with proven drift detection techniques for evaluation purposes.
While there are several established supervised drift detection methods such as Drift Detection Method~(DDM), Early Drift Detection Method~(EDDM), Adaptive Windowing~(ADWIN), and DDE, we focus on unsupervised methods due to their greater applicability in real-world scenarios~\cite{Gama2004,baena2006early,TimeChanging2007,maciel2015lightweight}.

To ensure reproducibility and practical application, we prioritize methods with publicly available source code and datasets.
Despite recent advances in unsupervised drift detection, obtaining experimental source code and datasets remain a significant challenge. 
Given our limited resources, implementing multiple algorithms from scratch is not feasible and may not yield consistent results due to differences in dependencies and optimizations.
%Although our algorithm can be used on top of any method, for our experiments, 
We select two algorithms that meet our criteria of having open-source code and public datasets: the Discriminative Drift Detector (D3) and the One Class Drift Detector (OCDD). 
The specific adaptations of these algorithms for our plugin method are detailed in~\autoref{subsec:d3} and~\autoref{subsec:ocdd}. 
The datasets used by the authors consist of real-world and artificial datasets from Losing et al.\cite{losing2016knn}. We use nineteen datasets, including both real-world and artificial datasets.

\subsection{Adapting the approach to D3}
\label{subsec:d3}
Discriminative Drift Detector~(D3)~works by obtaining a window $W$ of size~$w(1+\rho)$ of continuous data samples, from which two sub-windows of size $w$ and $w*\rho$ are considered~\cite{Omer2019}. 
One of the sliding windows, $W_{curr}$ of size $w$ represents the current distribution~($D_{curr}$), and the second sliding window represents the new distribution~($D_{next}$) of size $w*\rho$, and we represent this sub window as $W_{next}$ in this work. 
The underlying theory here is that if a classifier can distinguish between $W_{curr}$ and $W_{next}$, there is a separation between the two sub-windows, and that means a change in underlying distribution, which is a concept drift.

%For implementing~\SUDS~in D3, we need a separate function as the D3 algorithm predicts almost always the majority class due to the imbalance in data, while we need individual samples to retrain the classifier.
To implement~\SUDS~in D3, we take both the sub-windows~$W_{curr}$~and~$W_{next}$.
Since~$W_{curr}$~is larger in default parameter settings, we subsample~$W_{curr}$ so that $|W_{curr}| = |W_{next}|$.
%The majority class prediction is due to the imbalance between new data points and old data points~(10 vs. 100 in default parameters), which we alleviate by majority sub-sampling.
%Therefore, we formulate a separate algorithm for selecting samples as given in~\autoref{alg:D3_SUDS}. The algorithm is similar to the one of D3, and the only difference is that we subsample the $W_{curr}$ to equalize the number of samples under the windows $W_{curr}$ and $W_{next}$. 
A classifier, logistic regression in this case, is trained on the both~$W_{curr}$~and~$W_{next}$ as label 0 and label 1 respectively, and predict confidence scores for the whole window $W$. 
To train for the new distribution~$D_{next}$, we take the data points with the highest confidence scores for label 1.
%We choose the predicted data points with the highest confidence score of the new distribution. 
%We set the number of data points to be $|W_{next}|$, which in default parameter settings is ten data points. 
We then train the Hoeffding tree on these data points instead of simply taking the $W_{next}$ as done in D3.
We use the Hoeffding tree as the class predictor as in the original D3 and OCDD algorithms.

\begin{algorithm}
    \caption{Algorithm to extract homogeneous samples for D3}
    \begin{algorithmic}
    \Ensure $|W|=Window\_Size*(1+\rho)$
    \State $N\gets Window\_Size * \rho$
    \State $W_{curr} \gets Subsample(\{W_1, W_2,..W_{Window\_Size}\}$
    \State $W_{next} \in \{W_{Window\_Size+1} ...\}$
    \State $S\gets 0^{|W_{curr}| + |W_{next}|}$
    \State $S_i\gets1, 1 \leq i \leq |W_{curr}|$
    \State $clf\gets Train~LinearRegressor(W, S)$
    \State $samples\gets$ Top~|N| predictions~of~0~class
    \end{algorithmic}  
    \label{alg:D3_SUDS}
\end{algorithm}
%\vspace*{-1cm}
\subsection{Adapting the approach to OCDD} 
\label{subsec:ocdd}

One-Class Drift Detector is based on a One class SVM classifier performing the drift detection~\cite{Goezueacik2020}. 
OCDD works by taking a window, considering it as the current distribution, and then searching for outliers. 
The algorithm signals a drift once the number of outliers reaches a threshold. 
This detection is under the assumption that when the data distribution changes, the number of outliers detected will increase.

To incorporate our retraining strategy, we do the following: we fit a new classifier with the data contained in the new context sub-window.
We train another One-Class Support Vector Machine(SVM) classifier with only the data points marked as outliers.
We then predict on the whole window using the trained One-Class SVM. 
The theory is that we need homogeneous samples for the retraining, and the outliers should be part of the new distribution. 
We then decide based on the predictions of the old and new classifiers as given in~\autoref{alg:OCDD_SUDS}. 
The samples need to be in distribution based on the classifier trained on the new incoming data distribution and out of distribution for the classifier trained on the old or previous data distribution.
%We explain the \SUDS algorithm for OCDD in~\autoref{alg:OCDD_SUDS}. 
%The current Window in OCDD contains data points the algorithm marked as outliers. 
%Instead of taking the last samples, we train the Hoeffding Tree Classifier on the intersection of all the data points predicted as in distribution by the One-Class SVM and the outliers present within the window. 
%This selection ensures that the OCDD algorithm predicts the trained data as out of distribution and as in distribution samples by the One-Class SVM trained only on the out of distribution data.
We require homogeneous samples to train the Hoeffding tree and this is achieved by selecting the common data samples that are predicted as outliers according to the classifier trained on the previous data distribution and as indistribution according to the classifier trained on the new incoming data distribution.
\begin{algorithm}
    \caption{Algorithm to extract homogeneous samples for OCDD}
    \label{alg:OCDD_SUDS}
    \begin{algorithmic}
    \State $W\gets~Current Window under consideration$
    \State $W_{outliers}\gets~Outliers~from~W$
    \State $train~OneClassSVM(Outliers)$\Comment{clf is an outlier detection algorithm like One Class SVM}
    \State $samples\gets~predict\_clf(W)$\Comment{Gets all the in-distribution samples}
    \State $features\gets W_{outliers}\cap samples$ \Comment{features is the set of homogeneous samples}
    \State $TrainModel(features)$\Comment{Retrain using the homogeneous samples}
    \end{algorithmic}
\end{algorithm}
%\vspace*{-1cm}
\subsection{Harmonized Annotated Data Accuracy Metric(\metric)}
Most evaluation of drift detection methods focuses on accuracy metrics, often neglecting the crucial aspect that is the amount of labeled data required for effective model retraining~\cite{Goezueacik2020,Omer2019,gama2014survey}.
This oversight presents a significant challenge, as data labeling is expensive and can add noise~\cite{rasmussen2022challenge,yordanova2018challenges,caselli2020hatebert,wendlinger2021methods,spasic2020clinical}. 
In real-world applications, the usability of a machine learning system depends on both efficiency and effectiveness. 
Although there are a few other custom metrics for drift detection, none of them address the data annotation requirement\cite{seeliger2017detecting,ji2017effect}.
%Sato et al. specifically addresses this problem by looking into the different metrics such as Change Point(CP), Error Tolerance(ET) and so forth\cite{sato2021survey}.
%None of the metrics address the need to take the amount of training data into consideration which in a real-world scenario is one of the most expensive input.
Thus, there is a pressing need for a more holistic approach that balances model performance with the practical constraints of data annotation.

There is research in this direction such as sample size estimation for classifier accuracy but this area mainly focuses on static data distributions where data collection is time-consuming, expensive, or both~\cite{balki2019sample}.%,beleites2013sample}. 
%When dealing with concept drift in real-world data, it is essential to evaluate a method's performance not only based on raw metrics but also on the amount of training data required. 
%Reduced training data directly correlates with reduced effort and cost needed to obtain the ground truth data.
%It also reduces the time required to obtain the dataset to retrain the model for prediction. 
%In certain domains, such as intrusion detection, credit card fraud detection, and astrophysics, the creation of ground truth data necessitates the expertise of domain specialists. 
%This process is often expensive and labor-intensive, particularly when labeling large volumes of data points. 
%By optimizing the amount of training data needed, we can significantly alleviate these burdens, making the retraining process more efficient and cost-effective~\cite{fellicious2024driftgan}.
Considering all these factors, we introduce a metric~\longmetric(\metric)~to evaluate the performance of the classifiers based on the training data requirements and the effectiveness, accuracy in this case. 
%Our goal is to ensure that the metric balances the amount of annotated data with the model's performance.
We propose the metric as a harmonic mean based on the Pythagorean means of the performance metric~(e.g. Accuracy)~and the amount of unannotated samples in the dataset.
\metric~is calculated as follows.
\begin{equation}
    \frac{1}{\metric} = \frac{1}{2} \times \left( \frac{1}{\psi} + \frac{1}{\epsilon} \right) \\
    \Rightarrow \metric = 2 \times \frac{\psi \times \epsilon}{ \psi + \epsilon}
    \label{eqn:metric}
\end{equation}
where,
\begin{conditions}
    \psi &  the performance metric \\
    \epsilon &  $1-(|Annotated~Samples|/|Dataset|).$
\end{conditions}

We integrate $\epsilon$ as the percentage of unannotated samples instead of the raw number of samples to remove the influence of dataset size.
%Using the raw number of penalizes larger datasets as they will likely to contain more drifts. 
%The smaller dataset size will skew the metric towards smaller datasets, which tend to have fewer drifts and require much smaller annotated training samples. 
 
The higher the value, the better the model's overall performance on the dataset. 
Performance metrics such as accuracy, precision, recall, and f1-score are in [0, 1], and realistically, the percentage of samples will be (0, 1] and thus puts the score also in [0, 1). 
The metric, therefore, balances the raw performance and the amount of annotated samples. 
All our methods are publicly available\footnote{\url{https://github.com/cfellicious/SUDS/}}. 
%The repository includes the scripts for running the experiments and preparing the output data and the output files of the scripts.
\section{Results}
\label{sec:results}
For our comparisons, we required algorithms that used open datasets and publicly available code. 
Even though there are many recent advances in concept drift detection methods, such as Student Teacher concept for Unsupervised Drift Detection(STUDD)\cite{cerqueira2023studd}, we were not able to obtain the source code and hyperparameters necessary to implement and test it ourselves.
%Therefore, we selected two algorithms in the domain of unsupervised drift detection to run our modifications, and they are the One Class Drift Detection(OCDD) and Discriminative Drift Detection(D3) methods.
We also want to reduce the influence the random factors on the results and run our experiments multiple times with the default hyperparameters provided by the authors.
%We run the algorithms we selected for our experiment, along with their SUDS modifications twenty-five times with the same hyperparameters provided by authors for the original algorithms as well as the~SUDS~modifications. 
The accuracy metrics are computed based on the Interleaved Test-Then-Train method adopted by most drift detection algorithms including D3 and OCCD~\cite{Survey2013}
For D3, the hyperparameter values are $w=100$, $\rho=0.1$ and $\tau=0.7$, where $w$ is the window size, $\rho$ is the percentage of $w$ considered as the new window, and $\tau$ is the Area Under Curve(AUC)as explained by Gözüaçik et al.~\cite{Omer2019}.
For OCDD, there are two hyperparameters: the window size~$w$ and $\rho$ denote the percentage of the window that could be outliers. The authors use $w=250$ and $\rho=0.3$~\cite{Goezueacik2020}.
The~\SUDS~modifications of D3 and OCDD use the same parameters given by the authors.
%We compare the original D3 and OCDD algorithms and our modifications by accuracy, and ~\autoref{tbl:accuracy_comparison} presents the results. 
We chose accuracy as the metric, as the authors of the original papers used the same metric for comparison with existing methods. 
\autoref{tbl:accuracy_comparison}~presents the results of our experiments.
~\autoref{tbl:accuracy_comparison} shows that SUDS modifications perform similarly to their counterparts across most datasets but poorly on artificial datasets\footnote{\url{https://github.com/vlosing/driftDatasets}}. Notably, in 'InterchangingRBF,' D3 with~\SUDS~shows over 20\% difference compared to D3, and OCDD with~\SUDS~nearly 72\% compared to its non-SUDS version. We found that this discrepancy is due to abrupt drifts favoring the acquisition of the last few data points at drift points, as seen in D3 and OCDD~\cite{losing2016knn}.
Similar patterns appear in 'Moving Squares,' though less severe. In 'SEA big' and 'SEA stream',~OCDD(\SUDS)~performs best, attributed to 10\% noise complicating boundary detection. The~\SUDS~modifications excel with datasets resembling real-world data, whereas exceptionally clean datasets with only abrupt drifts do not align well with our method. In such rare cases, relying on the last few samples for training the new model is proven to be more effective.
%\vspace*{-1cm}
\begin{table}
    \centering
    \caption{Comparison of~\metric(~\autoref{eqn:metric})~values for D3 and OCDD along with respective SUDS modifications. The best result across all four methods is in bold, while the best in each method~(D3 and OCDD) is underlined.}
    \begin{tabular}{|c|c|c||c|c|}
    \hline
    Dataset & D3 & D3(SUDS)& OCDD & OCDD(SUDS)\\
    \hline
    Airlines & 0.7258 & \underline{0.7538} & 0.0016 & \textbf{\underline{0.7680}}\\
    Chessweka & 0.7927 & \textbf{\underline{0.7942}} & 0.6377 & \underline{0.6975} \\
    Covtype & 0.8862 & \textbf{\underline{0.9009}} & 0.0015 & \underline{0.4732}\\
    Electricity & \textbf{\underline{0.8857}} & 0.8844 & 0.0201 & \underline{0.8724}\\
    LUdata & 0.9266 & \textbf{\underline{0.9358}} & 0.3876 & \underline{0.8975}\\
    Outdoor & 0.7208 & \textbf{\underline{0.7572}} & 0.1984 & \underline{0.7458}\\
    Phishing & 0.9033 & \underline{0.9252} & 0.0829 & \textbf{\underline{0.9296}}\\
    Poker & \textbf{\underline{0.8272}} & 0.8203 & 0.0011 & \underline{0.7540}\\
    Rialto & \textbf{\underline{0.6656}} & 0.5957 & 0.0120 & \underline{0.5216}\\
    Spam & 0.8672 & \textbf{\underline{0.9069}} & 0.1302 & \underline{0.8903}\\
    Weather & 0.8093 & \textbf{\underline{0.8251}} & 0.0488 & \underline{0.5971}\\
    \makecell{Interchanging\\RBF} & \textbf{\underline{0.9328}} & 0.8074 & 0.0050 & \underline{0.4089}\\
    Mixed Drift & \textbf{\underline{0.6239}} & 0.6144 & 0.0015 & \underline{0.5646}\\
    Moving RBF & \textbf{\underline{0.6815}} & 0.6707 & 0.0050 & \underline{0.5081}\\
    Moving Squares & \textbf{\underline{0.8634}} & 0.7158 & 0.0050 & \underline{0.8039}\\
    \makecell{Rotating\\Hyperplane} & 0.8937 & \underline{0.9101} & 0.0050 & \textbf{\underline{0.9132}}\\
    Sea Big & 0.9064 & \underline{0.9102} & 0.0094 & \textbf{\underline{0.9373}}\\
    Sea Stream & 0.9078 & \underline{0.9096} & 0.0234 & \textbf{\underline{0.9309}}\\
    \makecell{Transient\\Chessboard} & \textbf{\underline{0.7206}} & 0.6623 & 0.0050 & \underline{0.5904}\\
    \hline
    \end{tabular}
    \label{tbl:lambda_comparison}
\end{table}
%\vspace*{-0.5cm}
%\vspace*{-8mm}
\begin{table}[!ht]
    \centering
    \caption{Accuracy comparison between D3, OCDD and D3, OCDD with SUDS. SUDS-modified methods are represented as D3(SUDS) and OCDD(SUDS) for D3 and OCDD respectively. The values for D3 and D3(SUDS are the mean over 25 runs with $\sigma$ in parentheses). The underlined method shows the best result with or without~\SUDS~modifications and the bold result shows the best result across all methods.}
    \begin{tabular}{|c|c|c||c|c|}
        \hline
        Dataset &  D3 & D3(SUDS) & OCDD & \makecell{OCDD\\(SUDS)}\\
        \hline
        Airlines & \makecell{60.40~(0.04)} & \makecell{\underline{61.57}~(0.07)} & 61.08 & \textbf{\underline{62.85}}\\
        Chessweka & \makecell{\textbf{\underline{67.98}}~(1.14)} & \makecell{67.83~(1.35)} & \underline{65.22} & \underline{65.22}\\
        Covtype & \makecell{\textbf{\underline{87.21}}~(0.02)} & \makecell{82.70~(0.66)} & \underline{84.70} & 84.04\\
        Electricity & \makecell{\textbf{\underline{86.51}}~(0.10)} & \makecell{80.60~(0.67)} & \underline{83.69} & 79.42\\
        LUdata & \makecell{88.43~(1.86)} & \makecell{\underline{89.68}~(2.44)} & 85.77& \textbf{\underline{90.73}}\\
        Outdoor & \makecell{60.00~(0.00)} & \makecell{\underline{61.65}~(1.57)} & 60.29 & \textbf{\underline{63.20}}\\
        Phishing & \makecell{86.45~(0.22)} & \makecell{\underline{88.04}~(0.34)} & \textbf{\underline{90.56}} & 89.77\\
        Poker & \makecell{\textbf{\underline{76.08}}~(0.04)} & \makecell{71.04~(0.32)} & \underline{74.62} & 72.02\\
        Rialto & \makecell{\underline{50.57}~(0.38)} & \makecell{42.63~(2.05)} & \textbf{\underline{66.68}} & 48.45\\
        Spam & \makecell{82.90~(0.21)} & \makecell{\underline{84.83}~(1.10)} & 85.43 & \textbf{\underline{87.04}}\\
        Weather & \makecell{\textbf{\underline{72.20}}~(0.12)} & \makecell{71.33~(0.44)} & \underline{71.66} & 71.23\\
        \makecell{Interchanging\\RBF} & \makecell{\underline{88.66}~(0.39)} & \makecell{67.94~(6.07)}& \textbf{\underline{97.02}} & 25.70\\
        Mixed Drift & \makecell{\textbf{\underline{45.39}}~(0.62)} & \makecell{44.35~(2.55)} & \underline{41.64} & 39.41\\
        Moving RBF & \makecell{\textbf{\underline{52.22}}~(0.17)} & \makecell{50.70~(0.91)} & \underline{51.96} & 34.07\\
        \makecell{Moving~Squares} & \makecell{\underline{76.64}~(0.78)} & \makecell{55.97\\(1.79)} & \textbf{\underline{95.89}} & 68.72\\
        \makecell{Rotating\\Hyperplane} & \makecell{82.98~(0.10)} & \makecell{\textbf{\underline{84.58}}~(0.18)} & 82.61 & \underline{84.07}\\
        Sea Big & \makecell{\underline{84.37}~(0.14)} & \makecell{84.10~(0.27)} & 83.17 & \textbf{\underline{88.31}}\\
        Sea Stream & \makecell{\underline{84.54}~(0.18)} & \makecell{84.06~(0.33)} & 83.16 & \textbf{\underline{87.35}}\\
        \makecell{Transient\\Chessboard} & \makecell{\underline{57.43}~(0.04)} & \makecell{49.75~(2.03)}& \textbf{\underline{57.61}} & 49.90\\
        \hline
    \end{tabular}
    \label{tbl:accuracy_comparison}
\end{table}
%\vspace*{-0.5cm}
%We initially intended to rank the original and modified approaches using performance positions across datasets to assess the methods' effectiveness. 
%However, relying solely on algorithm ranks offers a limited perspective. The problem lies in the discrete nature of ranking. 
We could compare our method against the original methods by ranking but ranking algorithms does not provide significant insights in our case due to the discrete nature of ranking.
Analyzing the performance differences between these methods would yield better insights beyond mere rankings.
%Therefore, we compute the average difference between a method and its~\SUDS~based counterpart. This value is computed from~\autoref{tbl:accuracy_comparison}.
Therefore, we compute the average difference of a method based on~\autoref{eqn_avg_diff}.
% shows the computation of the average difference.
\begin{equation}
    Avg~Diff_{Method} = \frac{1}{N} * \sum_{1}^{N} Max_{i} - Method_{i}
    \label{eqn_avg_diff}
\end{equation}
\begin{conditions}
    Method&Method$\in$\{D3,D3(SUDS),OCDD,OCDD(SUDS)\} \\
    Max_{i}&The best accuracy for Dataset~i \\
    Method_{i}&Accuracy for Dataset~i with Method
\end{conditions}
%\vspace*{-5mm}
\begin{table}
    \centering
    \caption{Average difference of percentage points to the best performing method for each dataset.}
    \begin{tabular}{|c|c|c|c|c|}
        \hline
         & D3 & D3(SUDS) & OCDD & \makecell{OCDD\\(SUDS)} \\
         \hline
         All datasets & 03.60 & 07.16 & 04.95 & 08.84 \\
         Real-world datasets & 02.94 & 04.48 & 07.16 & 03.37 \\
        \hline
    \end{tabular}
    \label{tbl:average_difference}
\end{table}
%\vspace*{-0.5cm}
%From~\autoref{eqn_avg_diff}, we understand that the lower scores are better overall and looking at these results, we see that D3 performs the best with only a 3.6\% average difference in accuracy to the best performing method of each dataset.
From~\autoref{eqn_avg_diff}~we know that the average difference will be in~$[0,1]$~with 0 being the best and 1 being the worst.
Looking at the scores, we see that D3 performs the best with only~3.6\%~average difference in accuracy to the best performing method of each dataset.
Here, the~\SUDS~modifications perform the worst with third and fourth places respectively. 
A deeper inspection reveals that this is due to the abrupt drifts present in the artificial datasets that causes the performance to drop.
We validate this assumption by looking at only the real world datasets where the~\SUDS~modifications are close to the best performing method overall which is D3.
The OCDD method drops places even though it requires the most training data.
Overall~D3(\SUDS)~performs at a deficiency of approximately~1.5\%, the method requires only 20\% of the annotated data when compared to D3.

We also look at the difference between the original algorithms and their~\SUDS~counterparts as well.
In this case, the average difference in accuracy between D3 and its~\SUDS~modification is 3.56\% with D3 exhibiting better performance while that of OCDD and corresponding~\SUDS~modification is 7.01\%.
When evaluating only real-world datasets, the difference shrinks to~1.53\%~between D3 and D3(\SUDS), with the gap between OCDD and OCDD(\SUDS)~reducing to 1.61\%.
When we put this into the amount of annotated data required as well from~\autoref{tbl:percentage_comparison}, we get a better picture.
Although we have to accept an average performance drop of less than 2\%, we require less than 20\% of the annotated data of the corresponding algorithm.
This is a huge factor for real-world applications where a slight drop in performance might be acceptable for a less time expensive method in terms of annotation time and its cost.

We can explain this reduced requirement for annotated data by examining the number of detected drifts in~\autoref{tbl:drift_count_comparison}.
It significantly affects the training data needed in real-world scenarios and also has an impact on the retraining time of the classifier. 
%With the default hyperparameters, D3 requires ten samples to retrain per detected drift while OCDD needs seventy-five.
%D3 typically requires ten samples per detected drift, while OCDD needs seventy-five samples per drift, both with default hyperparameters. 
%D3 with~\SUDS~modification also requires ten samples per drift. 
%However, OCDD(\SUDS) can vary, given its 250-sample window and the selection of samples belonging only to the new distribution.
%~\autoref{tbl:drift_count_comparison} shows the training data required for each method and the number of drifts detected. 
\begin{table*}[!ht]
    \centering
    \caption{Comparison of mean annotated data (25 runs) required between D3, OCDD and with corresponding SUDS modifications represented and detected drifts in parentheses. D3 requires ten samples per detected drift while OCDD needs seventy five samples per drift. For OCDD~(SUDS)~it may vary as the selected samples is not constant at every detected drift.}% For D3 and D3(SUDS), the results are the mean over 25 runs. The total number of instances in each dataset are given in |Dataset|. Lower numbers are better.}
    \begin{tabular}{|c|c|c|c|c|c|}
        \hline
        Dataset &  D3 & D3(SUDS) & OCDD & OCDD(SUDS) & |Dataset|\\
        \hline
        Airlines & 48,910(4891) & 15,151(1515) & 538,950(7186) & \textbf{7,048(179)} & 539,383 \\
        Chessweka & 25(2) &  \textbf{21(2)} & 75(1) & 126(1) & 503 \\
        Covtype & 57,641(5764) & \textbf{6,213(621)} & 580575(7741) & 389,669(5663) & 581,012\\
        Electricity & 4,202(420) & \textbf{924(92)} & 44,850(598) & 1463(35) & 45312\\
        LUdata & 51(5) & \textbf{41(4)} & 1,425(19) & 213(5) & 1901\\
        Outdoor & 390(39) & \textbf{76(8)} & 3,525(47) & 362(7) & 4000\\
        Phishing & 601(60) & \textbf{278(28)} & 10,575(141) & 399(10) & 11,055\\
        Poker & 77,653(7765) & \textbf{24,576(2458)} & 828750(11,050) & 173,171(3953) & 829,201\\
        Rialto & 2,184(218) & \textbf{927(93)} & 81,750(1090) & 35,794(624) & 82,250\\
        Spam & 566(57) & \textbf{161(16)} & 5,775(77) & 552(11) & 6,213\\
        Weather & 1,480(148) & \textbf{390(39)} & 17,700(236) & 8,826(137) & 18,159\\
        \makecell{Interchanging\\RBF} & 3,172(317) & 1,030(103)& 199,500(2660) & \textbf{135(3)} & 200,000\\
        Mixed Drift & 1,480(148) & \textbf{204(20)} & 599,550(7994) & 2,832(62) & 600,000\\
        Moving RBF & 3,859(386) & 1,900(190) & 199,500(2660) & \textbf{210(5)} & 200,000\\
        Moving Squares & 2,312(231) & \textbf{1,444(144)} & 199,500(2660) & 6,348(166) & 200,000\\
        \makecell{Rotating\\Hyperplane} & 6,313(631) & 3,005(300) & 199,500(2660) & \textbf{145(3)} & 200,000\\
        Sea Big & 2,069(207) & 829(83) & 99,525(1327) & \textbf{138(3)} & 100,000\\
        Sea Stream & 791(79) & 356(36) & 39,525(527) & \textbf{151(3)} & 40,000\\
        \makecell{Transient\\Chessboard} & 6,581(658) & \textbf{1,955(195)} & 199,500(2660) & 55,435(956) & 200,000\\
        \hline
    \end{tabular}
    \label{tbl:drift_count_comparison}
\end{table*}
%\vspace*{-10mm}
We see that our modifications D3(SUDS) and OCDD(SUDS) require the least amount of training data for every dataset, as shown in~\autoref{tbl:drift_count_comparison}. 
If we look at a percentage-wise comparison of annotated data required, OCDD requires the most~(over 90\%), while D3 requires considerably less. 
Introducing our modifications further reduces the amount of annotated data required to the original methods by approximately~80\%.
OCDD(\SUDS)~detects fewer drifts for the artificial datasets likely due to the absence of noise in some datasets. 
However, detecting fewer drifts only sometimes means worse performance.
For instance, in the "Rotating Hyperplane," "Sea Big," and "Sea Stream" datasets, OCDD(SUDS) requires approximately 150 annotated instances but either outperforms the best approach or is on par.% with  in "Sea Big" and "Sea Stream," and nearly as well as the best performer in "Rotating Hyperplane."

These examples demonstrate that an improved data sampling strategy can enhance performance while reducing the effort required for data labeling.
We show the two different views with~\autoref{tbl:accuracy_comparison} and~\autoref{tbl:drift_count_comparison}. 
%While achieving supreme model performance is desirable, it's impractical if it requires annotating nearly the entire dataset as seen with OCDD.
The~\metric~metric we introduce in~\autoref{eqn:metric}~and detailed in~\autoref{tbl:lambda_comparison}~balances performance and annotated data requirements.
Therefore, using our method provides a better understanding on how an algorithm performs in a real-world scenario.
Our modifications outperform existing methods in eleven of nineteen datasets. 
Notably, our methods struggle in artificial datasets lacking noise or with abrupt drifts. 
However, their performance improves notably when noise is introduced, as seen in the SEA concepts synthetic dataset with 10\% noise added.
Beyond raw performance, our formulated metric  in~\autoref{tbl:lambda_comparison}~considers training data volume alongside performance.
~\SUDS~outperforms original algorithms concerning annotated data for drift events. 
OCDD predicts more drifts than D3, demanding more annotated data and retraining. 
Despite OCDD's superior raw performance, the increased labelling effort often outweighs its performance advantage.
%\vspace*{-5mm}
\begin{table}[!ht]
    \centering
    \caption{Average percentage of dataset required as annotated training data.}
    \begin{tabular}{|c|c|c|}
         \hline
         Method & Annotated \% of Dataset  & Std. Dev\\
         \hline
         D3 & 05.03 & 03.49 \\
         D3(SUDS) & 01.68 & 01.04\\
         OCDD & 92.59 & 19.78 \\
         OCDD(SUDS) & 14.45 & 19.64\\
         \hline
    \end{tabular}
    \label{tbl:percentage_comparison}
\end{table}
%\vspace*{-5mm}
We can see this from~\autoref{tbl:percentage_comparison} where our method incorporated into D3 requires the least amount of training data(01.68\%). Here, we do not consider the labels required by the Interleaved-test-then-train method\cite{gama2014survey}. The amount of training data required by D3(SUDS), our modification to D3, is approximately one-third the amount of annotated data required by the original D3. Compared to OCDD, our modification~OCDD(SUDS) requires one-sixth of the annotated data required by OCDD. 

\subsection{Hyperparameter Search}

\iffalse
\begin{figure*}
    \centering
    \includegraphics[width=0.7\textwidth]{figures/results/D3_real.png}
    \caption{Average difference in~\metric~for real-world datasets between D3(SUDS) and D3 for different hyperparameter combinations. Compared to~\autoref{fig:d3_hyperparam} where both real-world and synthetic datasets are considered, here we consider only the real-world datasets. A positive value indicates that the SUDS modification is better on average across the datasets than the D3 for the same hyperparameters. There are no negative values indicating that~\SUDS~performs on average better than D3 in real-world conditions.}
    \label{fig:d3_hyperparam_realworld}
\end{figure*}
\fi
\begin{figure*}
    \centering
    \includegraphics[width=0.5\textwidth]{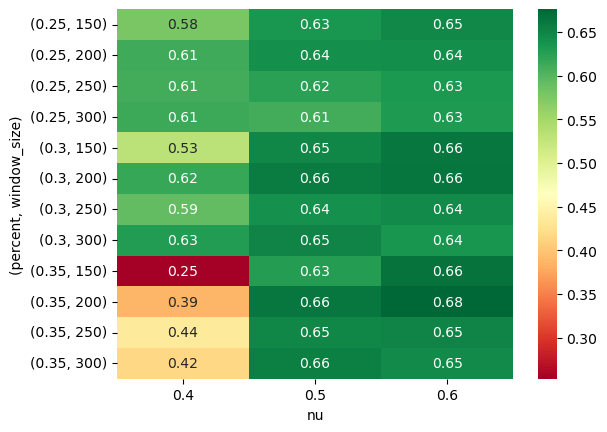}
    \caption{Average difference in~\metric~for real-world datasets between OCDD with and without~\SUDS~for different hyperparameter combinations. We see that our~\SUDS~modifications perform better overall as there are no negative values. A negative value indicates that the original algorithm was better.}
    \label{fig:OCDD_hyperparam}
\end{figure*}
\iffalse
\begin{figure*}
    \centering
    \begin{subfigure}{0.45\textwidth}
    \includegraphics[width=\textwidth]{figures/results/D3.png}
    \caption{Average difference in~\metric~between D3(SUDS) and D3 for different hyperparameter combinations.}
    \label{fig:d3_hyperparam}
    \end{subfigure}
    \begin{subfigure}{0.45\textwidth}
    \includegraphics[width=\textwidth]{figures/results/OCDD.png}
    \caption{Average difference in~\metric~between OCDD(SUDS) and OCDD for different hyperparameter combinations. Here, our SUDS modification performs better than OCDD for all hyperparameter combinations.}
    \label{fig:OCDD_hyperparam}
    \end{subfigure}
    \caption{Comparison between D3 and OCDD along with their corresponding~\SUDS~modifications. A positive value means that on average SUDS modification is better than the corresponding algorithm.}
    \label{fig:hyperparameter_search_both}
\end{figure*}
\fi
\begin{figure*}
    \centering
    \begin{subfigure}{0.48\textwidth}
    \includegraphics[width=\textwidth]{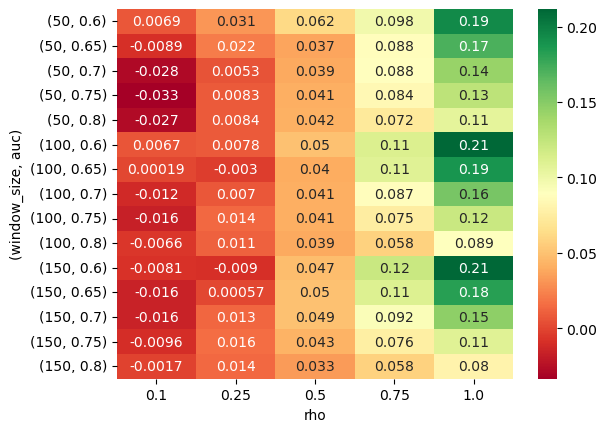}
    \caption{Average difference in~\metric~between D3(SUDS) and D3 across all datasets.}
    \label{fig:d3_hyperparam}
    \end{subfigure}
    \begin{subfigure}{0.48\textwidth}
    \includegraphics[width=\textwidth]{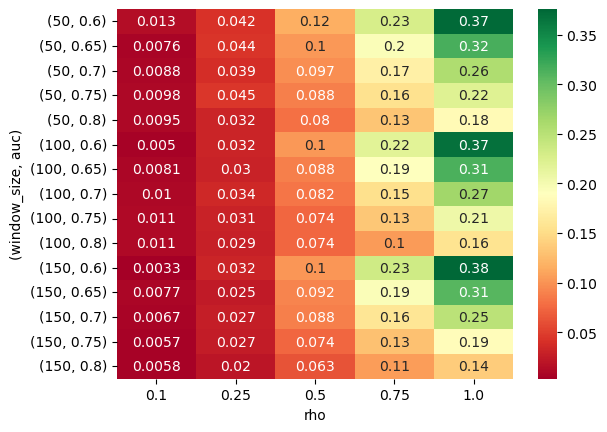}
    \caption{Average difference in~\metric~between D3(SUDS) and D3 across \emph{all real-world datasets}.}
    \label{fig:d3_hyperparam_realworld}
    \end{subfigure}
    \caption{Comparison of~\metric~between D3 and~D3(\SUDS)~algorithms across all datasets and across only real-world datasets, along with their corresponding~\SUDS~modifications. A positive value means that on average SUDS modification is better than the corresponding algorithm. We see that D3(\SUDS)~performs better for real-world datasets as there are only positive values in~\autoref{fig:d3_hyperparam_realworld}.}
    \label{fig:hyperparameter_search_both}
\end{figure*}
We further evaluate the performance of~\SUDS~across different hyperparameter combinations. For the D3 algorithm, the authors identify three hyperparameters which are:
\begin{itemize}
    \item $w$ is the window size considered as the old distribution
    \item $\rho$ is the fraction of the window size considered as the new distribution
    \item AUC is the area under the curve acting as the threshold to signal a drift
\end{itemize}

We consider $w$ = [50, \textbf{100}, 150], $\rho$ = [\textbf{0.1}, 0.25, 0.5, 0.75, 1.0], and $AUC$ = [0.6, 0.65, \textbf{0.7}, 0.75, 0.8] as the range of hyperparameters for D3 with the bold value indicating the default values.
For the window size~($w$), we evaluate both smaller and larger windows compared to the default value. 
Parameter $\rho$ includes five different values to assess whether a larger sample size for the new distribution is beneficial overall. 
Additionally, we examine different threshold values to determine if raising the threshold and thus reducing the number of detected drifts improves the overall algorithm performance.
We run the same hyperparameter combination three times and average it out to minimize the effect of outliers resulting in 225 runs for each dataset of the nineteen datasets. 
We then compute the~\metric~for each hyperparameter combination based on the average accuracy and average training data required. The average of~\metric~is computed across all datasets for both D3 and D3 with~\SUDS~modification. 
We then plot this difference, where positive values indicate better performance by D3 with~\SUDS, and negative values indicate better performance by the original D3. 
%We present the results of the hyperparameter study in~\autoref{fig:d3_hyperparam}.

From~\autoref{fig:d3_hyperparam}, our algorithm performs worse than the D3 algorithm only for the smallest~$\rho$~value~(0.1). 
%The degradation could be the result of having too few samples under the new distribution compared to the old distribution.
The~$\rho$~parameter decides the percentage of the window to be considered as the new distribution to check against the samples in the window.
Having too few samples could cause the sampling mechanism to perform worse as it is the case with most machine learning models.
Another reason could be that our model performs worse on the artificial datasets and this contributes to lowering the mean. 
To verify whether artificial datasets primarily contribute to the degradation of the~\SUDS~algorithm, we tabulate the results for only the real-world datasets.

From~\autoref{fig:d3_hyperparam_realworld}, the results show an improvement. All values are positive, indicating that our method performs better on average when compared to D3 w.r.t real world datasets. 
This could be due to the absence of sudden drifts which help in a better sampling strategy overall.
Another observation is that~\SUDS~performs better with a larger number of samples from the newer distribution.

For OCDD, there are only three hyperparameters,
\begin{itemize}
    \item $\nu$ parameter to the One-Class SVM that defines "the upper bound on the fraction of training errors and a lower bound of the fraction of support vectors"~\cite{scikit-learn}.
    \item $w$~is the size of the window under consideration
    \item $\rho$~is the threshold for drift detection
\end{itemize}
We consider $w$~=~[150, 200, \textbf{250}, 300], $\rho$~=~[0.25, \textbf{0.3}, 0.35] and $\nu$~=~[0.4, \textbf{0.5}, 0.6] as the parameter values for the hyperparameter search with the bold values indicating the default parameter values for the algorithm.
From~\autoref{fig:OCDD_hyperparam}~, we see that our method performs better overall for all hyperparameter combinations.
%\vspace*{-1cm}
\section{Conclusion}
\label{sec:conclusion}
\iffalse
This paper focuses on the challenges of concept drift in 
unsupervised learning and proposes a novel approach to improve 
an algorithm's adaptation to newer drifts. The proposed approaches are validated and evaluated through extensive experiments. 
The experimental results substantiate the effectiveness of our approach, demonstrating that the strategic acquisition of homogeneous data points for retraining reduces the need for extensive annotated training data, all while maintaining performance levels on par with conventional methods.
Furthermore, our work introduces a novel metric that reflects real-world applicability by amalgamating annotated data quantity with model performance, fostering a more practical assessment framework. 
Collectively, these findings underscore the potential of our approach to streamline and enhance the adaptability of unsupervised learning algorithms in dynamic real-world scenarios, offering both improved efficiency and performance.
\fi
We address the challenges of concept drift in unsupervised learning by proposing a novel approach to enhance an algorithm's adaptability to changing data distributions. 
We validate our approach through extensive experiments, which demonstrate its effectiveness. The results show that the strategic acquisition of homogeneous data points for retraining reduces the need for extensive annotated training data while maintaining performance levels comparable to conventional methods.
Additionally, we introduce a new metric that combines the quantity of annotated data with model performance, providing a more practical assessment framework for real-world applicability. 
These findings highlight the potential of our approach to improve the efficiency and adaptability of unsupervised learning algorithms in dynamic environments, offering both enhanced performance and efficiency.
\bibliographystyle{bibliography/IEEEtran}
\bibliography{bibliography/bibliography}

\end{document}